\documentclass[10pt,twocolumn,letterpaper]{article}

\usepackage{cvpr}              

\usepackage{CJK}

\usepackage[utf8]{inputenc} 
\usepackage[T1]{fontenc}    
\usepackage{hyperref}       
\usepackage{url}            
\usepackage{booktabs}       
\usepackage{amsfonts}       
\usepackage{nicefrac}       
\usepackage{microtype}      
\usepackage{xcolor}         
\usepackage{graphicx}
\usepackage{amsmath}
\usepackage{wrapfig} 
\usepackage{multirow}
\usepackage{pifont}   
\usepackage{xcolor}  
\usepackage{fontawesome}
\usepackage{graphicx}  

\usepackage[normalem]{ulem} 
\definecolor{blackpink}{rgb}{0.83, 0.19, 0.79}

\newcommand{\with}{\textcolor{green}{\ding{51}}}

\newcommand{\without}{\textcolor{red}{\ding{55}}}
\newcommand{\para}{\vspace{0.25cm}}

\title{LSceneLLM: Enhancing Large 3D Scene Understanding \\ Using Adaptive Visual Preferences}

\author{
    Hongyan Zhi\textsuperscript{\rm 1}\thanks{Equal contribution. Email: hoyard1212@gmail.com} ~~ 
    Peihao Chen\textsuperscript{\rm 2}\footnotemark[1] ~~ 
    Junyan Li\textsuperscript{\rm 4}\footnotemark[1] ~~  
    Shuailei Ma\textsuperscript{\rm 3} ~~
    Xinyu Sun\textsuperscript{\rm 1} ~~\\
    Tianhang Xiang\textsuperscript{\rm 1} ~~
    Yinjie Lei\textsuperscript{\rm 7} ~~
    Mingkui Tan\textsuperscript{\rm 1 \rm 6}\thanks{Corresponding author. Email: mingkuitan@scut.edu.cn} ~
    Chuang Gan\textsuperscript{\rm 4 \rm 5} \\
    \textsuperscript{\scriptsize{\rm 1}}\small{South China University of Technology,}
    \textsuperscript{\scriptsize{\rm 2}}\small{Tencent Robotics X,}
    \textsuperscript{\scriptsize{\rm 3}}\small{Northeastern University,} \\
    \textsuperscript{\scriptsize{\rm 4}}\small{UMass Amherst,}
    \textsuperscript{\rm 5}\small{MIT-IBM Watson AI Lab,}
    \textsuperscript{\scriptsize{\rm 6}}\small{Pazhou Laboratory,}
    \textsuperscript{\scriptsize{\rm 7}}\small{Sichuan University}\\
}

\begin{document}
\begin{CJK*}{UTF8}{gbsn}

\maketitle

\begin{abstract}
Research on 3D Vision-Language Models (3D-VLMs) is gaining increasing attention, which is crucial for developing embodied AI within 3D scenes, such as visual navigation and embodied question answering. Due to the high density of visual features, especially in large 3D scenes, accurately locating task-relevant visual information is challenging. 
Existing works attempt to segment all objects
and consider their features as scene representations. However, these task-agnostic object features include much redundant information and missing details for the task-relevant area.
To tackle these problems, we propose \textbf{LSceneLLM}, an adaptive framework that automatically identifies task-relevant areas by leveraging LLM's visual preference for different tasks, followed by a plug-and-play scene magnifier module to capture fine-grained details in focused areas.
Specifically, a dense token selector examines the attention map of LLM to identify visual preferences for the instruction input. It then magnifies fine-grained details of the focusing area. An adaptive self-attention module is leveraged to fuse the coarse-grained and selected fine-grained visual information.
To comprehensively evaluate the large scene understanding ability of 3D-VLMs, we further introduce a cross-room understanding benchmark, \textbf{XR-Scene}, which contains a series of large scene understanding tasks including XR-QA, XR-EmbodiedPlanning, and XR-SceneCaption. 
Experiments show that our method surpasses existing methods on both large scene understanding and existing scene understanding benchmarks. Plunging our scene magnifier module into the existing 3D-VLMs also brings significant improvement. 
Code and data are available at  \url{https://github.com/Hoyyyaard/LSceneLLM}
\end{abstract}

\begin{figure}[t]
    \centering
    \includegraphics[width=1\linewidth]{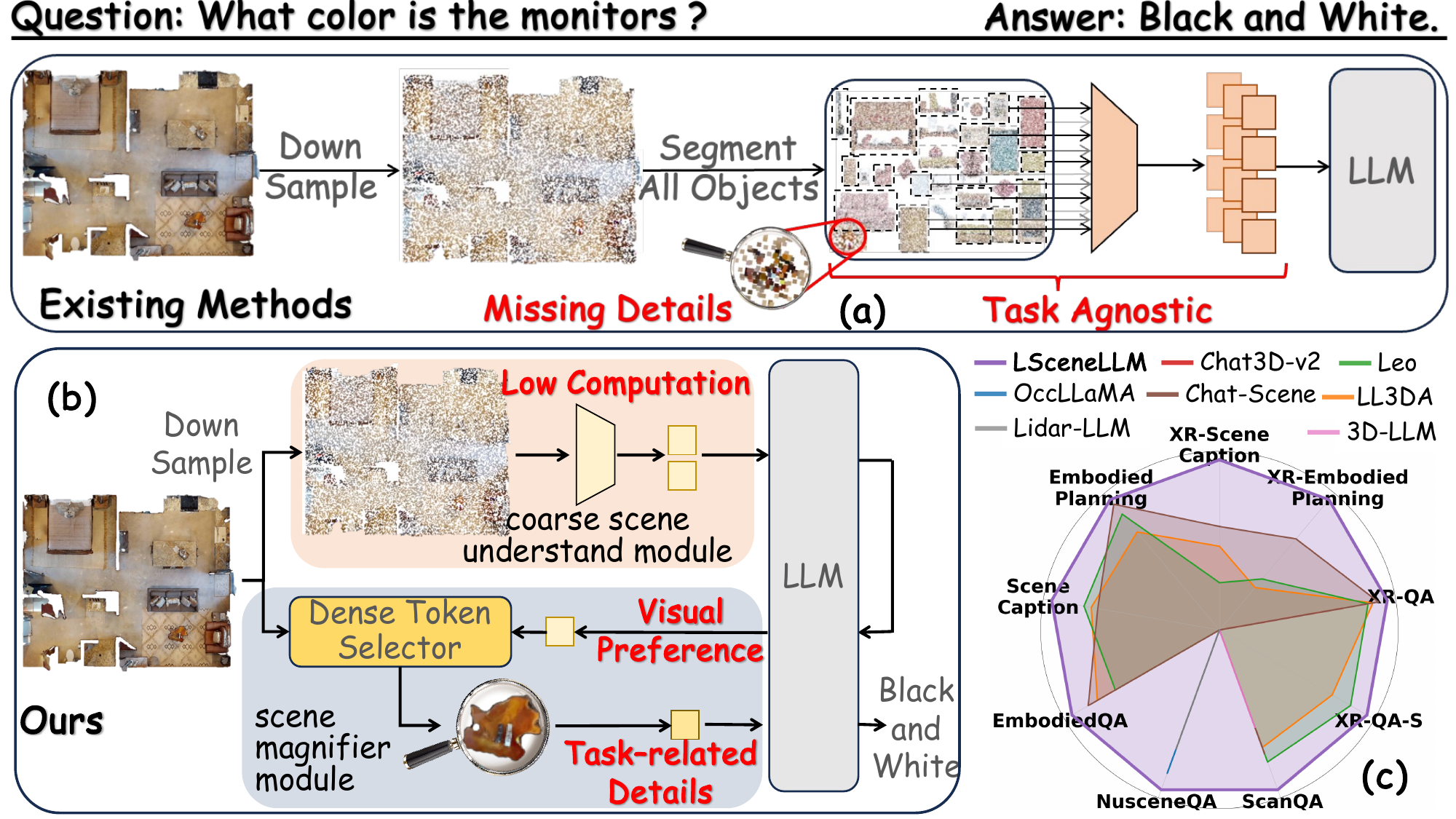}
    \caption{\textbf{We propose LSceneLLM, a novel framework for adaptive large 3D scene understanding.} (a) Existing methods struggle to locate
    task-relevant visual information when facing large scenes. (b) We are committed to precisely identifying fine-grain task-related visual features through adaptive scene modeling. (c) Our method outperforms existing approaches across various benchmarks.}
    \label{fig:teaser}
\end{figure}

\section{Introduction}
3D scene understanding is essential for various tasks, including robotic manipulation~\cite{zhen20243d,huang2023voxposer}, navigation~\cite{sun2024fgprompt}, and embodied long-horizon planning~\cite{chen20232,krantz2020beyond}. With the rapid development of large language models (LLMs)~\cite{bai2023qwen,chiang2023vicuna}, researchers are increasingly focusing on leveraging LLMs' impressive reasoning and summarization capabilities to enhance the understanding of 3D point clouds, building 3D Vision-Language Models (3D-VLMs)~\cite{chen2024ll3da,huang2023embodied,huang2023chat,hong20233d,xu2023pointllm,tang2024minigpt}.

Most efforts on 3D-VLMs have concentrated on object-level point cloud understanding~\cite{guo2023point,tang2024minigpt,xu2023pointllm}, while scene-level understanding remains underexplored due to the larger and more intricate nature of 3D environments compared to individual objects. 
To introduce the scene-level understanding 
ability to 3D-VLMs, existing works~\cite{chen2024ll3da,huang2023embodied,wang2023chat,huang2023chat,huang2024chatscenebridging3dscene} seek help from object detection~\cite{chen2023end} and instance segmentation~\cite{schult2023mask3d,zhang2023uni3d} technique to delineate all objects within a scene. All the object features extracted by an object-level 3D features extractor are considered to represent the scene-level 3D features, as illustrated in Fig.~\ref{fig:teaser}(a).
Despite the advancements, existing 3D-VLMs face two major limitations: 1) When facing large scenes, the scale of task-related visual information is significantly smaller than that of the entire scene. This disparity poses a significant challenge for existing 3D-VLMs in accurately focusing on pertinent visual information, as they require all segmented objects as input, which is task agnostic. 2) To balance the computational load, existing approaches utilize sparse object point clouds as input to capture object features within an entire scene, which leads to the loss of critical details related to small objects. 
LL3DA~\cite{chen2024ll3da} attempts to address the above issues by employing a Qformer~\cite{li2023blip} to query the task-relevant scene features. However, it fails to fully utilize the substantial reasoning capabilities of LLMs to assist in selecting task-relevant visual information and is hindered by the limitations of low-resolution scene features.
Moreover, Current 3D-VLM are predominantly benchmarked on the datasets that are annotated in ScanNet~\cite{dai2017scannet}, 3RScan~\cite{wald2019rio}, etc., which mainly consists of single-room~\cite{azuma2022scanqa,wald2019rio,chen2020scanrefer}. The understanding of large scenes such as multi-room scenarios remains underexplored. 

When faced with complex visual input, humans selectively focus on certain regions first, then search for information within those relevant areas~\cite{rao2002eye,wolfe2004attributes,treisman1980feature}. 
Mimicking the mechanisms humans use to process complex visual input, we posit that a model can first obtain visual preference within sparse scene representation, followed by a fine-grained analysis of focus areas. It's similar to how people, when reading a bulletin board, first focus on a specific topic before paying closer attention to particular details.

Inspired by the above motivation, we propose \textbf{LSceneLLM}, a 3D-VLM framework that contains a coarse scene understanding module and a scene magnifier module for adaptive modeling of large scenes.
As shown in Fig.~\ref{fig:teaser}(b), to obtain a preliminary understanding of the various areas of the scene, we utilize a scene encoder to encode the downsampled point cloud. Additionally, the scene magnifier module is proposed to identify the visual preference of LLM while extracting and fusing selected detailed visual information. Specifically, a dense token selector leverages the attention map of LLM to identify the visual preferences relevant to the instruction and then collects dense visual features from the areas of interest, which is then fused with coarse-grained scene information through the adaptive self-attention module, enabling large-scale scene understanding with limited point cloud input.
The scene magnifier module can be easily inserted into existing 3D-VLMs by replacing the corresponding self-attention module in the LLM. 

To provide a comprehensive evaluation for large scene understanding, we additionally propose a cross-room understanding benchmark \textbf{XR-Scene}, which includes XR-QA, XR-EmbodiedPlanning, and XR-SceneCaption. It has an average scene area size of 132 $m^{2}$, which is significantly larger than the 29 $m^{2}$ in ScanQA~\cite{azuma2022scanqa}.
Our experimental results demonstrate that LSceneLLM achieves state-of-the-art performance on a wide range of 3D tasks and benchmarks, including single-room scene benchmarks and large scene benchmarks, as shown in Fig.~\ref{fig:teaser}(c). Our contributions are summarized as follows:

\begin{itemize}
    \item We present LSceneLLM, a 3D-VLM framework that automatically identifies and magnifies detailed information in task-relevant areas. This helps the model accurately localize the important information within the large 3D scene.
    \item To comprehensively benchmark 3D-VLMs in large scene understanding, we present the XR-Scene, a collection of cross-room understanding tasks that includes question-answering, embodied planning, and scene caption, with an average scene area approximately four times larger than that of ScanQA, which offers a more challenging evaluation environment.
    \item Our approach consistently demonstrates superior performance in both indoor and outdoor large-scene understanding benchmarks. Integrating our scene magnifier module with existing 3D-VLMs also brings significant improvement.
\end{itemize}

\begin{figure*}[ht]
    \centering
    \includegraphics[width=1\textwidth]{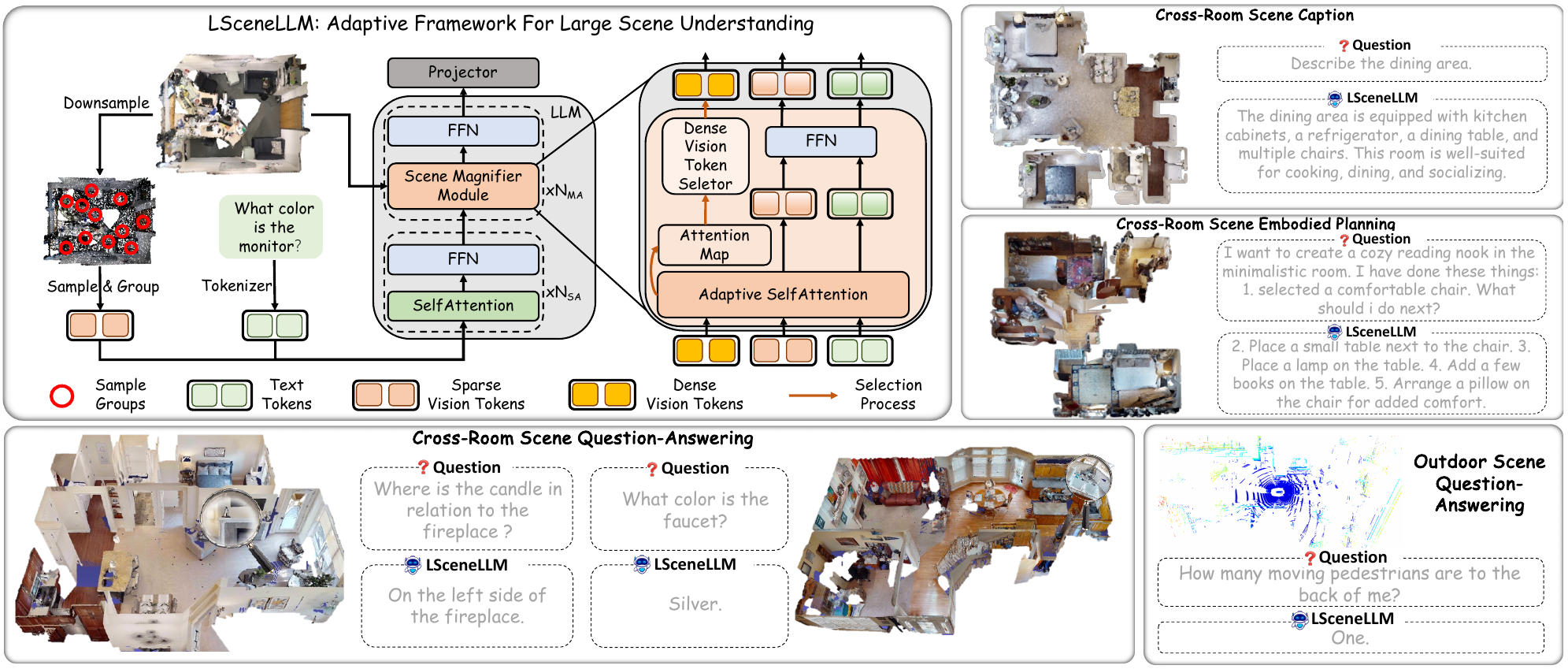}
    \caption{\textbf{An Overview of LSceneLLM.} LSceneLLM first perceives the scene through sparse vision tokens at the coarse level and then enhances regions of interest using dense vision tokens. Our method can effectively handle various visual language tasks in large scenes.}
    \label{fig:method}
\end{figure*}

\section{Related Work}

\subsection{3D Vision Language Models}
3D-VLMs have recently developed rapidly, thanks to the emergence of excellent work in LLMs. Researchers initially focused on understanding 3D object point clouds~\cite {xu2023pointllm,tang2024minigpt,guo2023point}. 
PointLLM~\cite{xu2023pointllm} first leverages the pretraining and instruction tuning paradigm to empower LLM to understand object point clouds. MiniGPT-3D~\cite{tang2024minigpt} introduced an efficient 3D-VLM that aligns 3D point clouds with LLMs by leveraging 2D priors from 2D-LLMs, capitalizing on the similarities between 2D and 3D visual information. Point-Bind~\cite{guo2023point} constructs a joint embedding space that integrates 3D and multimodal data, projecting 3D semantic features into LLMs to enhance their capacity for 3D object-level question answering.
Scene-level point clouds are inherently more complex than object-level point clouds due to the high density of visual features. This complexity poses a greater challenge for LLMs to understand scene-level point clouds effectively. Object-centric works~\cite{wang2023chat,huang2023embodied,huang2024chatscenebridging3dscene,huang2023chat} first leverage instance segmentation or object detection to extract all objects within the scene and model the spatial relationship between objects through a relation module. When facing large scenes, the scale of task-related visual information is significantly smaller than that of the objects features within the scene, as it involves querying a single object among hundreds of others. This challenges the task-agnostic visual feature construction process, making it difficult for the model to locate task-relevant information accurately.
A recent work, LL3DA~\cite{chen2024ll3da}, proposed using QFormer~\cite{li2023blip} to extract visual features related to task instructions. It employs a set of learnable queries to summarize the object features in the scene that are relevant to the task. However, this process occurs outside the LLM and is a shallow task-related information extraction process that demands a substantial amount of training data~\cite{yao2024deco}. We propose to leverage the LLM's powerful reasoning capabilities to select visual preferences based on different instructions.

\subsection{3D Understanding Benchmarks}
Early datasets for 3D understanding, such as NYUv2~\cite{silberman2011indoor} and SUN RGB-D~\cite{song2015sun}, comprise short RGBD sequences with low resolution and limited annotations. ScanNet~\cite{dai2017scannet} is the first dataset that provides 3D reconstructions and annotations at scale, including 1,201 and 312 different and complex indoor 3D scenes. Most existing indoor scene understanding benchmarks~\cite{chen2020scanrefer,azuma2022scanqa,achlioptas2020referit3d} are built from ScanNet. In addition, benchmarks~\cite{huang2023embodied} based on scenes like 3RScan~\cite{wald2019rio} and ARKitScenes~\cite{baruch2021arkitscenes} are also widely discussed. Despite the above advancements, existing 3D understanding benchmarks are largely limited to small scenes, such as single-room scenarios, and benchmarks for large scene understanding in cross-room and outdoor environments~\cite{qian2024nuscenes} remain underexplored. HM3D~\cite{ramakrishnan2021habitat} is the largest-ever dataset of 3D indoor spaces, consisting of 1,000 high-resolution 3D scans of building-scale residential, commercial, and civic spaces generated from real-world environments. We construct a large scene understanding benchmark based on HM3D for cross-room scenarios.

\begin{figure*}[ht]
    \centering
    \includegraphics[width=1\textwidth]{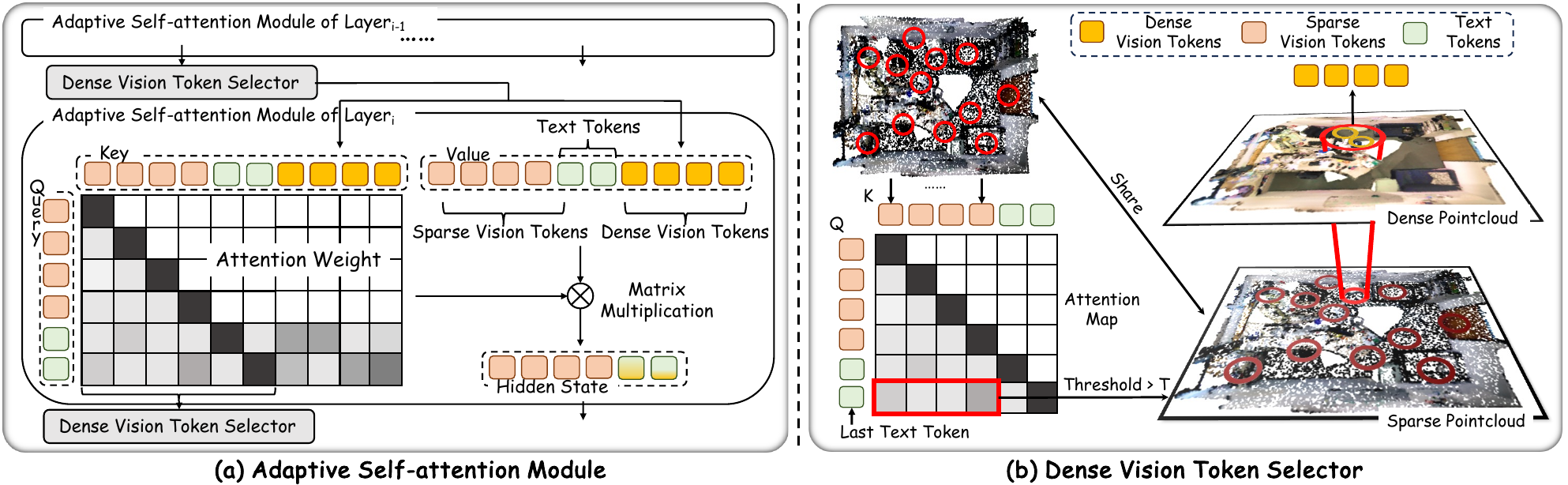}  
    \caption{\textbf{Illustration of Adaptive Self-attention Module and Dense Vision Token Selector.} We first obtain the focused regions by analyzing the attention map of LLM. Then we extract dense point cloud features from the region of interest and parse dense vision tokens through sampling and grouping operations. }  
    \label{fig:selector}  
\end{figure*}

\section{LSceneLLM: Adaptive Framework For Large Scene Understanding} 
When facing large scenes, to precisely identify task-relevant visual information within high-density visual contexts, we introduce \textbf{LSceneLLM}, which automatically
identifies task-relevant areas by leveraging LLM’s visual preference and then searching for the desired information within the focus areas. The adaptive framework consists of a coarse scene understanding module and a scene magnifier module, allowing for the comprehension of scenes from coarse-grained overviews to fine-grained details of significant regions. The scene magnifier module can be seamlessly integrated into most 3D-VLMs by simply replacing their corresponding self-attention modules in LLM.

\subsection{Overall Architecture}
As shown in Fig.~\ref{fig:method}, given the dense point cloud of a 3D scene, we first obtain dense point cloud features through a scene encoder~\cite{peng2023openscene}. These features are then down-sampled to sparse point cloud features. Through a coarse scene understanding module modified from SA module~\cite{qi2017pointnet++} that random sample points and groups a wide range of nearby features for each sample point, the sparse point cloud features are converted into sparse vision tokens, representing a coarse scene depiction. For fine-grain feature construction, we identify the center points of a preferred region and extract a specific number of point cloud features from the dense point cloud features in its vicinity. These local region point cloud features go through the SA module~\cite{qi2017pointnet++} to obtain the dense vision tokens. After receiving the visual features, we modify the LLM with our proposed scene magnifier module to achieve a fine-grained scene perception. Given an LLM with $N$ layers of self-attention, we replace the last $N_{MA}$ layers' self-attention modules with our scene magnifier module, while keeping the first $N_{SA}$ layers unchanged because the first few layers of attention tend to focus on the global information of the vision input. To reduce computational complexity, we input only the sparse vision tokens and text tokens into the first $N_{SA}$ layers. For the last $N_{MA}$ layers which utilize the scene magnifier module, we additionally incorporate selected dense vision tokens to enhance the model's understanding of regions of interest in greater detail. Specifically, the scene magnifier module comprises two sub-modules: a dense vision token selector and an adaptive self-attention mechanism. The dense vision token selector dynamically identifies and selects dense vision tokens for regions of interest guided by the attention map from the self-attention mechanism, rather than inputting all dense vision tokens. The adaptive self-attention mechanism integrates the information from the selected dense vision tokens into the original hidden states, thereby enriching the model's contextual understanding.

\subsection{Dense Vision Tokens Selector}
In an auto-regressive model, the prediction of the next token is contingent upon the hidden state of the last token processed. As shown in Fig.~\ref{fig:selector}(b), by analyzing the activation values of the attention map for the last token to the sparse vision tokens, we can identify specific visual information that the model focuses on when making predictions. Although sparse vision tokens provide limited information, we enhance the model's understanding by retrieving dense point cloud features in focused areas. We first extract the attention weights of the last text token to all sparse vision tokens and normalize them to a range of 0 to 255. Next, we select 10\% to 20\% sparse vision tokens whose weights exceed a specified threshold. For each selected sparse vision token, we identify the corresponding region in the scene and sample dense vision tokens from that region to provide richer visual information, thereby enhancing the model's understanding of the scene.

\begin{figure*}[ht]
    \centering
    \includegraphics[width=1\textwidth]{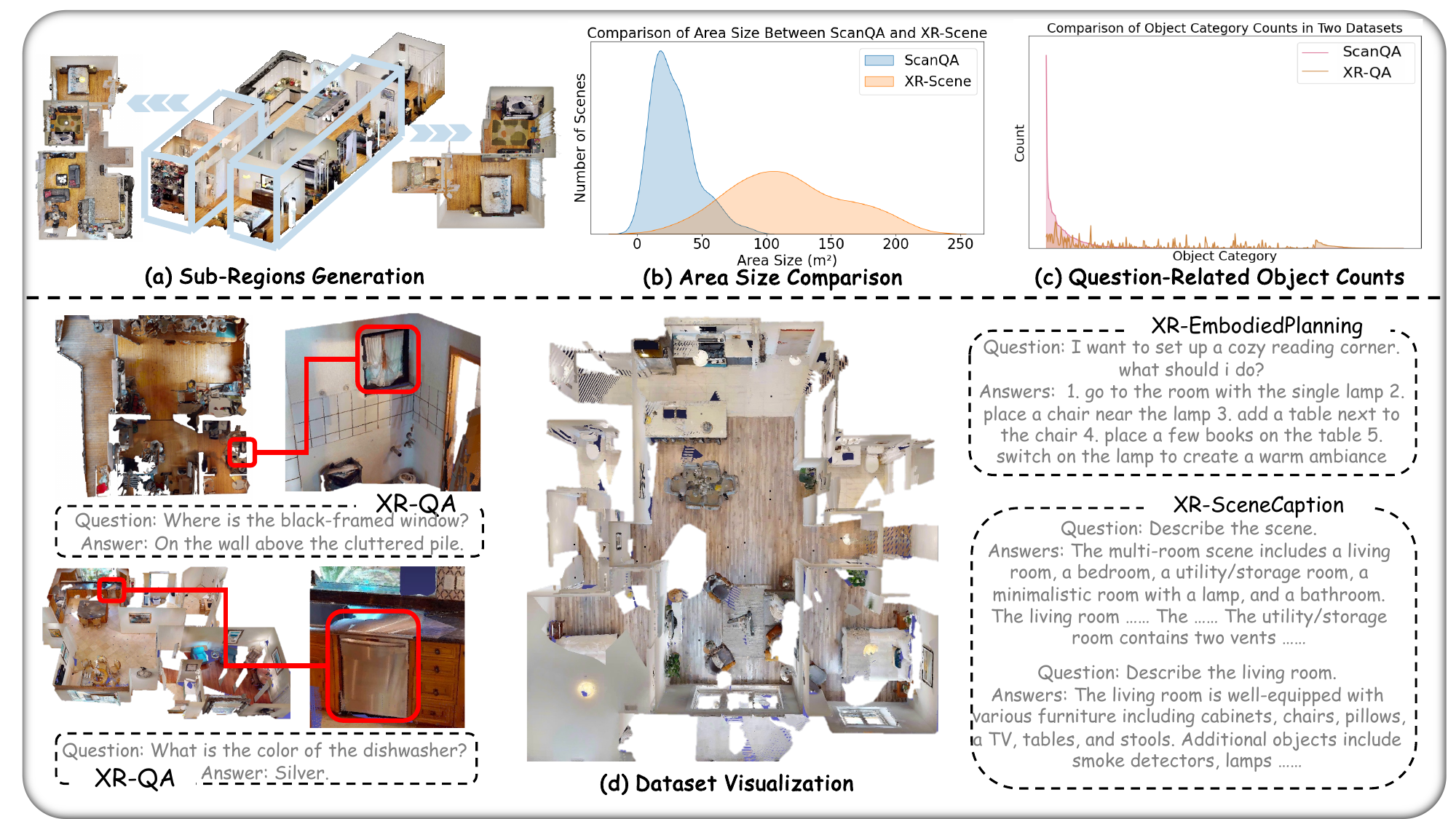}
    \caption{\textbf{Examples of dataset XR-Scene}. XR-Scene contains three cross-room scene benchmarks that comprehensively evaluate different understanding abilities.}
    \label{fig:dataset}
\end{figure*}

\subsection{Adaptive Self-attention Module}
The adaptive Self-attention Module is a crucial mechanism that integrates dense vision tokens $D$ with the hidden state $H$ containing text tokens and sparse vision tokens. As shown in Fig.~\ref{fig:selector}(a), this module takes dense vision tokens and the hidden state as input and outputs the fused hidden state. Unlike standard self-attention, our attention map captures interactions between the hidden state and dense vision tokens. We eliminate the attention map for this interaction component, and the remaining attention map is used to select dense visual tokens in the subsequent layer. It is important to note that each text token interacts solely with the selected dense vision tokens, which we achieve through the use of an attention mask. The calculation of the adaptive self-attention is summarized as follows:
\[
Q = \text{HW}_Q 
\]
\[
K_{\text{all}} = \text{Concat} (\text{HW}_K, \text{DW}_K') 
\]
\[
V_{\text{all}} = \text{Concat} (\text{HW}_V, \text{DW}_V') 
\]
\[
\text{Adaptive Self-attention} (H, D) = \text{softmax} \left( \frac{QK_{\text{all}}^T}{\sqrt{d_k}} \right) V_{\text{all}} 
\]
where \( W_Q, W_K, W_V, W'_K, W'_V \in \mathbb{R}^{D \times d} \) are learnable linear projection matrices. \( K_{\text{all}} \) and \( V_{\text{all}} \) are the key and value matrices that aggregate information from dense point cloud features.

\begin{table*}[]
\caption{3D large scene understanding results. All use Ll3da and XR-Scene data for training. $^*$ means do not identify the question-related objects for the model. $^\#$ means requiring images and point clouds as input.}
\label{tab:hd_results}
\centering
\resizebox{0.9\textwidth}{!}{%
\begin{tabular}{lccccccccc}
\toprule
\multirow{2}{*}{Methods} & \multicolumn{3}{c}{XR-QA}                         & \multicolumn{3}{c}{XR-SceneCaption}                                & \multicolumn{3}{c}{XR-EmbodiedPlanning}                            \\ \cmidrule(r){2-10} 
                         & CIDEr           & METEOR         & ROUGE          & CIDEr                & METEOR               & ROUGE                & CIDEr                & METEOR               & ROUGE                \\ \midrule
\multicolumn{10}{l}{\textbf{Zero-Shot}}  \\
Chat-Scene$^\#$~\cite{huang2024chatscenebridging3dscene}               &  69.55
& 26.63
& 10.06                & 0.01
& 5.94 & 1.52 & 32.64 & 20.71 & 10.26
 \\
Leo$^*$~\cite{huang2023embodied}                      & 55.40          & 22.71          & 6.96          & 0.02                 & 1.92                & 2.92               & 9.74              & 16.84                & 6.88                \\
Ll3da~\cite{chen2024ll3da}   & 24.78            & 12.66     & 5.31          & 0.12                & 8.71                & 5.14                & 7.02                & 15.21                & 7.17                \\
\midrule
\multicolumn{10}{l}{\textbf{Finetuning}}  \\
Chat-Scene$^\#$~\cite{huang2024chatscenebridging3dscene}               &  114.10
& 35.93
& 14.32                & 3.58
& 17.49 & 11.59 & 46.18 & 22.34 & 36.71
 \\
Leo$^*$~\cite{huang2023embodied}                      & 112.09          & 35.47          & 14.02          & 2.42                 & 15.96                & 10.25                & 39.45                & 18.99                & 33.31                \\
Ll3da~\cite{chen2024ll3da}   & 112.80                 & 36.94 & 18.68           & 3.22                 & 20.95                & 13.49                & 35.96                & 15.74                & 31.50                \\
\textbf{\textbf{LSceneLLM(Ours)} }               & \textbf{117.21} & \textbf{38.18} & \textbf{19.30} & \textbf{4.59}        & \textbf{23.43}                & \textbf{16.16}       & \textbf{63.08}       & \textbf{22.97}       & \textbf{36.96}       \\ \bottomrule
\end{tabular}
}
\end{table*}

\begin{table}[h]
\caption{3D question answering results on the ScanQA~\cite{azuma2022scanqa} validation dataset. $^*$ means do not identify the question-related objects for the model.}
\centering
\label{tab:scanqa}
\resizebox{0.9\linewidth}{!}{%
\begin{tabular}{lccccc}
\toprule
Method     & LLM        & Training Data & ROUGE          & METEOR         & CIDEr                    \\ \midrule
3D-VLP~\cite{jin2023context}      & -        & -              & 34.51          & 13.53          & 66.97                      \\
ScanQA~\cite{azuma2022scanqa}     & -        & -              & 33.33          & 13.14          & 64.86                       \\
Chat3D~\cite{wang2023chat}        &Vicuna-7b     & -             & 28.5           & 11.9           & 53.2                         \\
Chat3D-v2~\cite{huang2023chat}    & Vicuna-7b       & 204k             & 40.1           & 16.1           & 77.1               \\
3D-LLM~\cite{hong20233d}         & BLIP2-flanT5     & 675k          & 35.7           & 14.5           & 69.4            \\ 
SceneLLM~\cite{fu2024scene}  & Llama2-7b        & 690k          &   35.9        & 15.8          & 80.00              \\
Chat-Scene~\cite{huang2024chatscenebridging3dscene}  & Vicuna-7b & 145k & 37.79 & 15.94 & 77.75 \\
Leo$^*$~\cite{huang2023embodied}    & Vicuna-7b           & 1034k+145k           & 40.24
& 16.68 & 80.20   \\
Ll3da~\cite{chen2024ll3da}      & Opt-1.3b          & 145k          & 37.02          & 15.37          & 75.67               \\
Ll3da~\cite{chen2024ll3da}      & Llama2-7b        & 145k          & 38.31          & 15.91          & 79.08              \\        
\textbf{LSceneLLM(Ours)}   & Llama2-7b  & 145k          & \textbf{40.82}          & \textbf{17.95} & \textbf{88.24}    \\ 
\bottomrule
\end{tabular}%
}
\end{table}

\begin{table*}[]
\centering
\caption{3D question answering results on outdoor scene benchmark NuscenesQA~\cite{qian2024nuscenes}. * means downstream specialist model.}
\label{tab:nuscenes_qa}
\centering
\resizebox{0.9\textwidth}{!}{%
\begin{tabular}{lcccccccccccccccc}
\toprule
\multirow{2}{*}{Method}           & \multicolumn{3}{c}{Exist} & \multicolumn{3}{c}{Count} & \multicolumn{3}{c}{Object} & \multicolumn{3}{c}{Status} & \multicolumn{3}{c}{Comparison} & \multirow{2}{*}{Acc}  \\ \cmidrule(r){2-13}
                  & H0      & H1     & All    & H0      & H1     & All    & H0      & H1      & All    & H0      & H1      & All    & H0       & H1       & All  &                                                                         \\ \midrule
NuscenesQA*~\cite{qian2024nuscenes}        & 87.7    & 81.1   & 84.1   & 21.9    & 20.7   & 21.3   & 70.2    & 45.6    & 49.2   & 62.8    & 52.4    & 55.9   & 81.6     & 68.0     & 69.2     & 58.1                                                            \\
\midrule
LLaVA-Adaptaer-v2~\cite{gao2023llama} & 34.2    & 6.3    & 19.3   & 5.0     & 0.1    & 2.7    & 23.7    & 4.6     & 7.6    & 9.8     & 11.3    & 10.8   & 2.6      & 1.5      & 1.6      & 9.6                                                             \\
LLaVA~\cite{liu2024visual}             & 38.9    & 51.9   & 45.8   & 7.7     & 7.6    & 7.7    & 10.5    & 7.4     & 7.8    & 7.0     & 9.9     & 9.0    & 64.5     & 50.8     & 52.1     & 26.2                                                            \\
LidarLLM~\cite{yang2023lidar}          & 79.1    & 70.6   & 74.5   & 15.3    & 14.7   & 15.0   & 59.6    & 34.1    & 37.8   & 53.4    & 42.0    & 45.9   & 67.0     & 57.0     & 57.8     & 48.6                                                             \\
OccLLaMA~\cite{wei2024occllama}  & 80.6 & 79.3 & 79.9 & 18.6 & 19.1 & 18.9 & \textbf{64.9} & 39.0 & 42.8 & 48.0 & 49.6 & 49.1 & 80.6 & 63.7 & 65.2 & 53.4 \\
\textbf{LSceneLLM(Ours)}           & \textbf{86.4}            & \textbf{81.3}            & \textbf{83.6}            & 19.4            & \textbf{19.8}            & \textbf{19.6}            & 64.4            & \textbf{41.3}            & \textbf{44.8}            & \textbf{58.8}            & \textbf{51.2}            & \textbf{53.8}            & \textbf{81.0}            & \textbf{67.5}            & \textbf{68.7}            & \textbf{56.4}                                  \\ \bottomrule
\end{tabular}
}
\end{table*}

\begin{table*}[]
\caption{More results on the XR-QA validation dataset and challenge subset XR-QA-S. $^\#$ We re-implement Leo~\cite{huang2023embodied} and Ll3da~\cite{chen2024ll3da} keeping all other settings the same as ours to conduct a fair and further comparison.}
\label{tab:XR-QA}
\centering
\resizebox{0.9\textwidth}{!}{%
\begin{tabular}{lcclllll}
\toprule
\multirow{2}{*}{Method} & \multirow{2}{*}{Scene Magnifier Module}       & \multicolumn{3}{c}{XR-QA} & \multicolumn{3}{c}{XR-QA-S} \\ \cmidrule(r){3-8} 
                        &                                                              & ROUGE   & METEOR   & CIDEr   & ROUGE    & METEOR   & CIDEr   \\
\midrule
Leo$^{\#}$~\cite{huang2023embodied}     & \without              & 36.56   & 18.61    & 110.33  & 36.10    & 18.06         & 103.16   \\
Leo$^{\#}$~\cite{huang2023embodied}     & \with                                & 37.53$_\text{(+0.97)}$   & 19.00$_\text{(+0.39)}$    & 113.46$_\text{(+3.13)}$  & 36.88$_\text{(+0.77)}$    & 18.47$_\text{(+0.41)}$         & 107.56$_\text{(+5.29)}$   \\ 
\midrule
Ll3da$^\#$~\cite{chen2024ll3da}        & \without          & 37.19   & 18.51    & 111.35  & 36.04    & 17.61         & 95.65    \\
Ll3da$^\#$~\cite{chen2024ll3da} & \with                                & 37.85$_\text{(+0.65)}$   & 19.15$_\text{(+0.56)}$    & 115.79$_\text{(+4.44)}$  & 37.23$_\text{(+1.19)}$    & 18.60$_\text{(+0.99)}$         & 106.73$_\text{(+11.09)}$   \\ 
\midrule
LSceneLLM       & \without          & 36.58   & 18.65    & 109.92  & 35.47    & 17.91         & 97.57    \\
\textbf{LSceneLLM(Ours)}   & \with                                   & \textbf{38.18$_\text{(+1.60)}$}   & \textbf{19.30$_\text{(+0.65)}$}    & \textbf{117.21$_\text{(+7.29)}$}  & \textbf{38.15$_\text{(+2.68)}$}    & \textbf{18.69$_\text{(+0.78)}$}         & \textbf{109.42$_\text{(+11.85)}$}   \\ 
\bottomrule
\end{tabular}%
}
\end{table*}

\section{XR-Scene: Cross-Room Scene Understanding Benchmark}
Current benchmarks for 3D-VLMs primarily benchmark on the datasets that are annotated in ScanNet~\cite{dai2017scannet}, etc., which mainly consists of single-room scenes, such as ScanQA~\cite{azuma2022scanqa} and ScanRefer~\cite{chen2020scanrefer}. Benchmarking in cross-room scenes remains underexplored. Cross-room scenes exhibit higher spatial complexity and object diversity, posing a challenge for 3D-VLMs to comprehend complex scenes with limited scene point cloud inputs. To conduct a comprehensive evaluation of large scene understanding, we propose a cross-room understanding benchmark \textbf{XR-Scene}, which includes three tasks, XR-QA, XR-EmbodiedPlanning, and XR-SceneCaption, as shown in Fig.~\ref{fig:dataset}(d). 

We generate cross-room scenes from HM3D~\cite{ramakrishnan2021habitat} using ground-truth room positions, selecting the nearest $N$ rooms to form each scene. Object annotations from SceneVerse~\cite{jia2024sceneverse} are used to assist in the generation of different types of QA pairs. We prompt GPT-4o 
with annotations from each room and a top-down view of the whole scene to generate XR-Scene. \textbf{XR-QA} necessitates that 3D-VLM initially identifies the specific region within a vast scene to which the object associated with the query pertains. Furthermore, it must comprehend the relationships between this object and its surrounding entities to deliver precise responses. \textbf{XR-EmbodiedPlanning} requires 3D-VLM to comprehend complex inter-object and room relationships to produce subtasks based on high-level goals. Meanwhile, \textbf{XR-SceneCaption} challenges 3D-VLM to generate comprehensive scene descriptions and captions for specific rooms while inferring attributes based on present objects. Overall, these tasks demand advanced spatial reasoning and contextual awareness to navigate the intricacies of cross-room environments effectively. For more detailed information on dataset generation, please refer to Appendix~\ref{appendix:generation details}.

\subsection{Analysis of XR-Scene}
XR-Scene consists of more than 1000 unique scenes, with an average area of \textbf{132$m^2$}, significantly larger than the average scene area of \textbf{29$m^2$} in ScanQA~\cite{azuma2022scanqa}, as mentioned in Fig.~\ref{fig:dataset}(b). Moreover, the number of inquiries about objects in the scene of XR-QA is relatively balanced, whereas for ScanQA~\cite{azuma2022scanqa}, there is a heavy focus on asking numerous questions about chairs, and tables, resulting in a lack of diversity, as shown in Fig.~\ref{fig:dataset}(c). Furthermore, due to the visual density in large scenes, which often require down-sampling and can result in the loss of scene details, we selected a subset from XR-QA based on the bounding box sizes of the question-specific objects. This subset, \textbf{XR-QA-S}, has a bounding box size threshold of 0.05 $m^2$, equivalent to the size of a keyboard. This subset assesses the model's ability to comprehend fine-grained information in various environments.

\begin{figure*}[ht]
    \centering
    \includegraphics[width=1\textwidth]{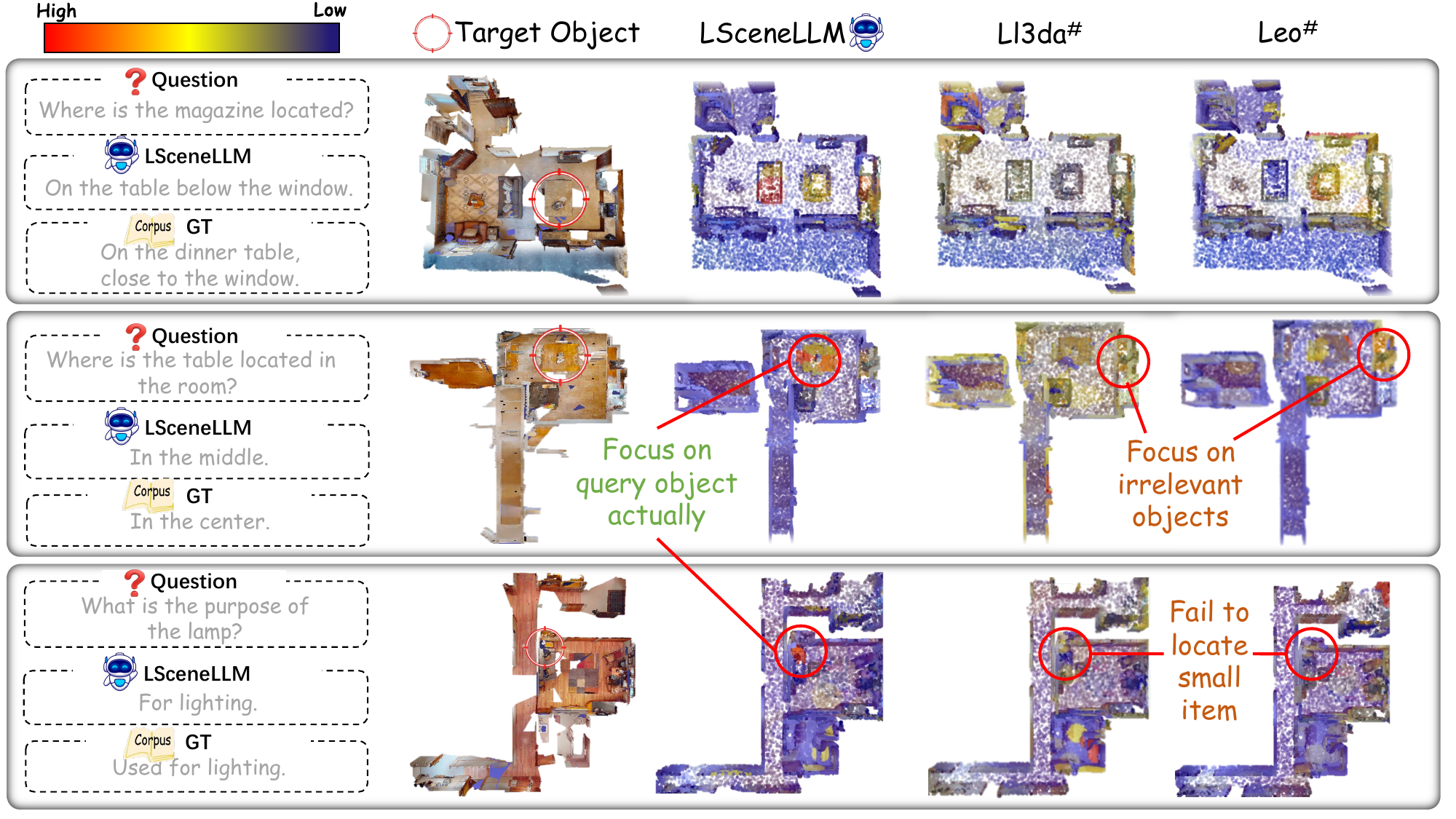}
    \caption{\textbf{Visualization of attention map of LLM.} Red represents high activation values, while blue represents low activation values.}
    \label{fig:vis_attn}
\end{figure*}

\begin{table}[]
\caption{Ablation studies. ATR: the activate token ratio of sparse vision tokens. $^\#$: do not use the scene magnifier module.}
\label{tab:ablation}
\centering
\resizebox{0.9\linewidth}{!}{%
\begin{tabular}{ccccc}
\toprule
\multicolumn{2}{c}{Parameter}                                  & ROUGE          & METEOR         & CIDEr           \\ \midrule
\multirow{3}{*}{Threshold}       & 96(AT: 10\%-20\%)           & \textbf{38.18} & \textbf{19.30} & \textbf{117.21} \\
                                 & 127(AT: 3\%-5\%)            & 37.89          & 19.26          & 115.92          \\
                                 & 64(AT: 40\%-50\%)           & 37.68          & 19.07          & 114.69          \\ \midrule
\multirow{3}{*}{Dense Token Num} & 2                           & 37.91          & 19.14          & 115.32          \\
                                 & 4                           & \textbf{38.18} & \textbf{19.30} & \textbf{117.21} \\
                                 & 6                           & 37.54          & 19.03          & 115.14          \\ \midrule
\multirow{2}{*}{Select Strategy} & Attention Map               & \textbf{38.18} & \textbf{19.30} & \textbf{117.21} \\
                                 & Random                      & 37.64          & 19.18          & 115.66          \\ \midrule
\multirow{3}{*}{Vision Token Num} & 512                         & 37.27          & 18.80          & 112.19          \\
                                 & 128                         & 36.58          & 18.65          & 109.92          \\
                                 & 128$^\#$ & \textbf{38.18} & \textbf{19.30} & \textbf{117.21} \\ \midrule
\end{tabular}%
}
\end{table}

\section{Experiment}
\subsection{Datasets, Metrics and Implementation Details}

\paragraph{Datasets.}
In this paper, we conduct experiments using 3D data from ScanNet~\cite{dai2017scannet} and HM3D~\cite{ramakrishnan2021habitat}. The annotated training data is sourced from ScanRefer~\cite{chen2020scanrefer}, Nr3D~\cite{achlioptas2020referit3d}, ScanQA~\cite{azuma2022scanqa}, the ScanNet subset of 3D-LLM~\cite{hong20233d} and our XR-Scene. The annotations cover various 3D tasks such as object captions, scene descriptions, scene question answering, and task planning. We also conduct experiments on the outdoor large scenes QA benchmark, Nuscenes-QA~\cite{qian2024nuscenes}, to validate whether our method effectively handles sparse and wide-ranging point clouds. Due to the huge gap between indoor and outdoor scenes, we tested with additional models, trained with Nuscenes-QA data.
\vspace{-0.3cm}

\paragraph{Metrics.}
Here, we adopt CiDEr~\cite{vedantam2015cider}, METEOR~\cite{banerjee2005meteor}, Rouge~\cite{lin2004rouge} and accuracy to evaluate the quality of the generated textual responses.
\vspace{-0.3cm}

\paragraph{Implementation Details.}
Following previous works on 3D vision language tasks~\cite{chen2024ll3da}, we randomly sample 40k point clouds from each 3D scene as the 3D visual input. We use the LLama2-7b~\cite{touvron2023llama} as our causal LLM backbone, which is fully fine-tuned to enable the model to attend to the areas of interest accurately. 
We adopt the AdamW~\cite{loshchilov2017decoupled} optimizer with a weight decay
of 0.1 and a learning rate decaying from 10$^{-5}$
to 10$^{-6}$ with a cosine annealing scheduler. 
In XR-Scene, We select Leo~\cite{huang2023embodied}, Chat-Scene~\cite{huang2024chatscenebridging3dscene} and Ll3da~\cite{chen2024ll3da} as the typical methods. We train all baselines with the same data in order to conduct a fair comparison.

\subsection{Comparison with SoTA Specialists}
\subsubsection{Indoor Scene Understanding} 
We benchmark existing methods on XR-Scene to evaluate the model's understanding of large scenes, as shown in Tab.~\ref{tab:hd_results}. We first evaluate existing methods in a zero-shot manner, and experiment results indicate that those models perform poorly without having undergone a learning process on large scene data. For question-answering task XR-QA, when facing high-density vision input, the performance of the Ll3da~\cite{chen2024ll3da} is poor due to the suboptimal process of extracting task-related visual information and detail loss in feature compressing process. Object-centric methods like Leo~\cite{huang2023embodied} and Chat-Scene~\cite{huang2024chatscenebridging3dscene} also perform poorly due to the overwhelming number of objects in the scene, which hinder the model's ability to focus on the relevant objects precisely. 
For XR-SceneCaption and XR-EmbodiedPlanning, which require a deep understanding of the relationships between regions (rooms) and objects to generate captions for specific areas or to complete high-level planning tasks within a region, existing methods yielded unsatisfactory results. This is because they modeled the scene only at the object level, neglecting the dependency information between regions and objects within the broader context of the scene. On the contrary, our proposed method, LSceneLLM, can identify task-relevant regions and facilitate a fine-grained understanding within focus areas. As a result, our approach outperforms existing methods across all three tasks by a large margin.

We also evaluate our method on ScanQA~\cite{azuma2022scanqa}, as shown in Tab.~\ref{tab:scanqa}. The experimental results indicate when facing small-scale scenes, our method can also achieve optimal performance with limited training data, further validating the effectiveness of the hierarchical scene understanding approach we proposed. More ScanNet understanding results can be found in Appendix~\ref{appendix:more_scannet_resules}.

\subsubsection{Outdoor Scene Understanding}
Outdoor scenes are larger in scale compared to indoor scenes, and the visual information is more sparse, which poses greater challenges for 3D-VLMs in understanding outdoor sparse point clouds.
We also conducted experiments on the NuscenesQA~\cite{qian2024nuscenes} benchmark to assess the model's capability in handling large outdoor scenes. Results in Tab.~\ref{tab:nuscenes_qa} show that our LSceneLLM achieves state-of-the-art performance across all generative methods without requiring multi-view image input or specialized framework design, with an increase of 3.0 in accuracy compared to the previous best method.

\subsection{More Insights into Fine-Grain Large Scene Understanding}
To reveal the capability of 3D-VLMs to understand small objects in scenes, we further conducted a more in-depth analysis of existing methods on XR-QA and XR-QA-S, as shown in Tab.~\ref{tab:XR-QA}. We re-implement Leo~\cite{huang2023embodied}
and Ll3da~\cite{chen2024ll3da} to keep all other settings the same as ours to conduct a fair comparison. The performance of existing methods on XR-QA-S has significantly declined compared to XR-QA.  This is due to the loss of details during downsampling required by object recognition techniques, making it difficult to perceive small objects in the scene accurately. Our proposed LSceneLLM can automatically complete fine-grained visual information and performs best on XR-QA-S, outperforming the existing methods on CIDEr by a large margin.

\subsection{Module Plug-and-Play Analysis}
To further verify that the scene magnifier module can be easily integrated into most existing 3D-VLM frameworks to enhance fine-grain understanding, we integrated it with two major architectures Leo~\cite{huang2023embodied} and Ll3da~\cite{chen2024ll3da}, as shown in Tab.~\ref{tab:XR-QA}. For the Ll3da framework, adding the scene magnifier module resulted in a 4.44 improvement in XR-QA and an 11.09 improvement in XR-QA-S. In comparison, the improvement was less significant compared to ours. We attribute this to the fact that the use of modules that compress visual information like Qformer~\cite{li2023blip} has resulted in the loss of visual details. Leo models the entire scene directly at the object level, so the improvement after adding the scene magnifier module is less compared to the other two frameworks due to the absence of the coarse scene understanding module.

\subsection{Ablation Studies}
We conduct several ablation studies, including the number of vision tokens, the dense vision token selection strategy, the selection threshold of attention value, and the number of dense vision tokens that interact with sparse vision tokens using the XR-QA dataset, as shown in Tab.~\ref{tab:ablation}. We experimentally found that using the attention map from LLM to identify the areas of interest and activating 10\%-20\% of these areas yields the best performance for our method. For more detailed information, please refer to Appendix~\ref{appendix:abl_study}.

\subsection{Qualitative Analysis}

\paragraph{Visualization of Attention Map}
We explore the areas the model focuses on when answering questions by visualizing the attention maps of the generated sequence of LLM to scene vision tokens, as shown in Fig.~\ref{fig:vis_attn}. Experiments were conducted using LSceneLLM and two other commonly used 3D-VLM frameworks Ll3da~\cite{chen2024ll3da} and Leo~\cite{huang2023embodied}. As illustrated by the example in the first row, when asked about the location of the magazine which is on the table in the scene, LSceneLLM accurately identifies the magazine's position and correctly focuses on the table mentioned in the response. The other two methods only roughly focus on the magazine's location and output the wrong answers. When asked about smaller objects (the third row), LSceneLLM is also able to accurately locate the position of the small objects, while the other two methods fail to do so.

\section{Conclusion}
In this paper, we investigate the paradigm of 3D-VLMs for large 3D scene understanding. To precisely locate task-related visual information, we propose an adaptive framework that automatically identifies task-relevant areas by leveraging LLM's visual preference when tackling different tasks, followed by a plug-and-play scene magnifier module to capture fine-grained details in focused areas. Experimental results demonstrate that our approach achieves significant performance improvements on large 3D scene understanding benchmarks. We further propose the XR-Scene cross-room understanding benchmark to complete the benchmarking process for 3D-VLMs in large-scale environments. We hope that our findings will inspire further advancements in the development of large 3D scene understanding methodologies.

\clearpage

{\small
\bibliographystyle{plain}
\bibliography{ref}
}

\clearpage

\appendix
\begin{leftline}
	{
		\LARGE{\textsc{Appendix}}
	}
\end{leftline}

\section{More Details on Generation of XR-Scene}

\label{appendix:generation details}
\begin{figure*}[h]
    \centering
    \includegraphics[width=1\textwidth]{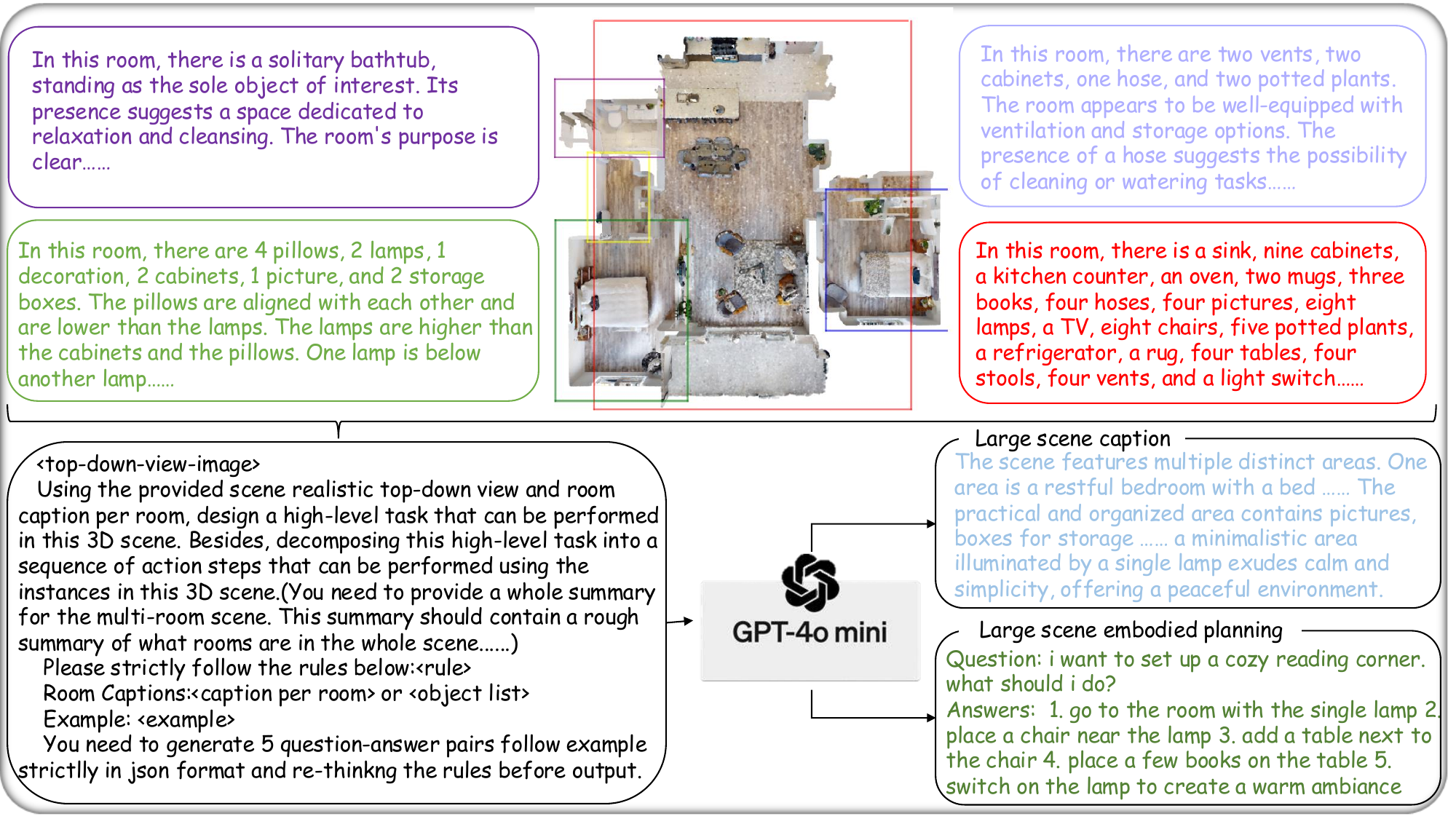}
    \caption{\textbf{Generation pipeline of 
 XR-SceneCaption and XR-EmbodiedPlanning}.}
    \label{fig:dataset_gen}
\end{figure*}

\label{appendix:viz_attn}
\begin{figure*}[h]
    \centering
    \includegraphics[width=1\textwidth]{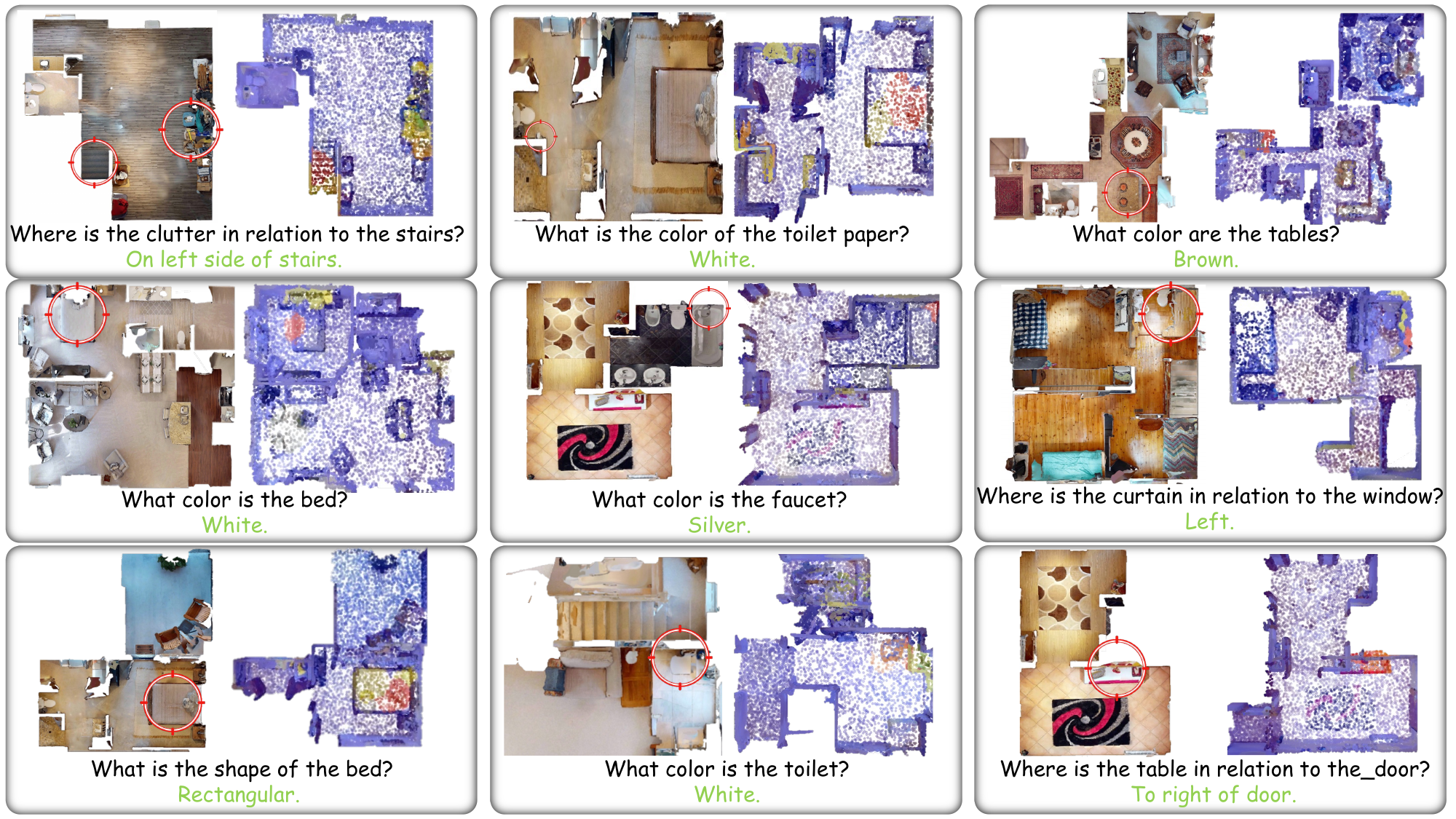}
    \caption{More Attention Visualization of LSceneLLM.}
    \label{fig:appx_viz_attn}
\end{figure*}

\para
\noindent\textbf{Generation of Cross-Room Scenes}
HM3D~\cite{ramakrishnan2021habitat} contains several cross-room, multi-floor 3D scenes. In SceneVerse~\cite{jia2024sceneverse}, annotations are generated for each room in the HM3D scenes, including object properties and spatial relationships with surrounding objects. As shown in Fig.~\ref{fig:dataset}(a). For a given scene, we leverage the ground-true central positions of each room in HM3D. We randomly sample one room and calculate the Euclidean distances between other rooms, the nearest $N$ rooms are selected to form a cross-room scene.

\para
\noindent\textbf{XR-QA Generation}
For each cross-room scene containing $N$ rooms, we retrieve object annotations from SceneVerse~\cite{jia2024sceneverse} for these $N$ rooms and \textbf{filter out objects that appear exactly once} in the scene to ensure uniqueness corresponding to the question. For each annotated object, we use GPT-4 to generate two types of questions: object properties and spatial relationships with surrounding objects based on the annotations.

\para
\noindent\textbf{XR-Planning and XR-EmbodiedPlanning Generation}
The embodied planning task requires the model to understand the objects in the scene and their specific locations. Given a high-level task, the model needs to use the objects in the scene to generate a series of subtasks. In contrast to single-room scenes, embodied planning in cross-room scenes is more complex for the model, as it needs to understand the relationships between objects and the rooms, not just the relationships between the objects themselves.

\noindent The scene captioning task requires the model to provide a general description of the current scene, including the relationships between objects and their attributes. In larger scenes, scene captioning demands a stronger spatial understanding of the model. The model not only needs to perceive the positional relationships between objects but also pay attention to the areas to which the objects belong. Our tasks will include generating captions for the entire large scene as well as requiring the model to caption only a specific room. Furthermore, the model needs to infer room attributes based on the objects present, making scene captioning in cross-scene Scenes more challenging than in single-room Scenes.

\noindent We generate the top-down view of the cross-room scene and use bounding boxes to specify that a certain annotation corresponds to a specific room. Follow Leo~\cite{huang2023embodied}, We use prompt engineering to guide GPT-4o in understanding the scene and generating scene captions and QA pairs for embodied planning. Additionally, we provide the model with a real RGB-rendered top-down view of the scene to further reduce model hallucinations, as shown in Fig.~\ref{fig:dataset_gen}.

\begin{table}
\centering
\caption{Ablation studies of selection threshold}
\resizebox{0.9\linewidth}{!}{%
\begin{tabular}{cccccc}
\toprule
Threshold & Activate Token Ratio & ROUGE  & METEOR & CIDEr  \\ \midrule
64        & 40\% - 50\%          & 37.68  & 19.07  & 114.69 \\
96        & 10\% - 20\%          & \textbf{38.18} & \textbf{19.30} & \textbf{117.21} \\
127       & 3\% - 5\%            & 37.89  & 19.26 & 115.92 \\ \bottomrule
\label{abl_select_thres}
\end{tabular}%
}
\end{table}
\begin{table}
\caption{Ablation studies of the number of vision tokens}
\centering
\resizebox{0.9\linewidth}{!}{%
\begin{tabular}{ccccc}
\toprule
Vision Token Num & Scene Magnifier Module & ROUGE  & METEOR & CIDEr  \\ \midrule
512   & \without       & 37.27        & 18.80       & 112.89        \\ 
128  & \without & 36.58 & 18.65 & 109.92 \\
128  & \with  & \textbf{38.18} & \textbf{19.30} & \textbf{117.21} \\
\bottomrule
\end{tabular}%
}
\label{abl_vision_num}
\end{table}

\begin{table}
\centering
\caption{Ablation studies of dense token}
\resizebox{0.6\linewidth}{!}{%
\begin{tabular}{cccc}
\toprule
Dense Token Num & ROUGE  & METEOR & CIDEr  \\ \midrule
2               & 37.91       & 19.14       & 115.32       \\
4               & \textbf{38.18} & \textbf{19.30} & \textbf{117.21} \\
6               & 37.54  & 19.03  & 115.14 \\ \bottomrule
\label{abl_dense_token_num}
\end{tabular}%
}
\end{table}

\begin{table}
\centering
\caption{Ablation studies of selection strategies}
\resizebox{0.6\linewidth}{!}{%
\begin{tabular}{cccc}
\toprule
Select Strategy & ROUGE  & METEOR & CIDEr  \\ \midrule
Attention Map   & \textbf{38.18} & \textbf{19.30} & \textbf{117.21} \\
Random          & 37.64        & 19.18       & 115.66        \\ \bottomrule
\end{tabular}%
}
\label{abl_strategy}
\end{table}

\begin{table*}[]
\centering
\caption{More 3D scene understanding results. $^*$ means do not identify the question-related objects for the model.}
\label{tab:other_results}
\small
\resizebox{0.9\textwidth}{!}{%
\begin{tabular}{lccccccccc}
\toprule
\multirow{2}{*}{Method} & \multicolumn{3}{c}{Scene Caption}               & \multicolumn{3}{c}{Embodied Planning}             & \multicolumn{3}{c}{Embodied QA}                   \\ \cmidrule(r){2-10} 
                        & ROUGE         & CIDEr          & METEOR         & ROUGE          & CIDEr           & METEOR         & ROUGE          & CIDEr           & METEOR         \\ \midrule
Leo*~\cite{huang2023embodied} & 1.80 & 20.84 & 13.29  & 46.40 & 204.78 & 19.86  & 30.89 & 86.14 & 18.81
 \\ 
Chat-Scene~\cite{huang2024chatscenebridging3dscene}  &\textbf{3.67} & 21.05 & 12.60 & 40.03
& 210.86 & 20.71 & 34.23 & 99.01 & 18.48 \\
Ll3da~\cite{chen2024ll3da}                   & 1.44          & \textbf{24.62} & 12.93          & 45.34          & 186.13          & 19.60          & 33.75               & 95.53                & 19.81               \\
\textbf{LSceneLLM(Ours)}           & 3.07 & 21.88          & \textbf{14.79} & \textbf{47.05} & \textbf{214.63} & \textbf{21.05} & \textbf{36.00} & \textbf{104.98} & \textbf{21.26}          \\ \bottomrule
\end{tabular}%
}
\end{table*}

\begin{table}[]
\centering
\small
\caption{Computational complexity results on XR-QA}
\label{tab:flops}
\resizebox{0.9\linewidth}{!}{%
\begin{tabular}{ccccc}
\toprule
Method    & Scene Magnifier Module & Vision Token Num & Flops & CIDEr  \\ \midrule
Leo       & \without                       & 200              & 6.55  & 110.33 \\
Ll3da     &  \without                      & 32               & 4.11  & 111.35 \\
LSceneLLM & \without                       & 128              & 5.3   & 109.92 \\
LSceneLLM &  \with                      & 128              & 6.33  & \textbf{117.21} \\ \bottomrule
\end{tabular}%
}
\end{table}

\section{Ablation Study}
\label{appendix:abl_study}

\para
\noindent\textbf{Selection Threshold of Attention Weight.}
We also explored the threshold for the confidence of text tokens to vision tokens in the attention map. We normalized the attention weight of a text token to all vision tokens to a range of 0-255. The experimental results show that the model performs best when the chosen threshold is 96, meaning 10\%-20\% of the vision tokens are selected to interact with the corresponding fine-grained scene features. If too many tokens are selected, the model cannot accurately focus on the local areas, while if too few tokens are selected, the fine-grained scene information provided is insufficient, offering limited help in understanding the scene, as shown in Tab.~\ref{abl_select_thres}.

\para
\noindent\textbf{Numbers of Dense Vision Token  Interact With Sparse Vision Token.}
This ablation experiment investigates the optimal number of dense vision tokens with which each sparse vision token should interact. We sample a certain number of point cloud features around the center point of the sparse vision token from the dense point cloud features and then aggregate them. As shown in Tab.~\ref{abl_dense_token_num}, using 4 dense vision tokens to represent the fine-grained features of a local region provides the greatest benefit to the model.

\para
\noindent\textbf{The number of Vision Tokens}
We first explored whether sampling more visual information from the environment would improve the model's performance. As shown in Tab.~\ref{abl_vision_num}, although using four times the number of vision tokens does lead to some performance improvement, the enhancement is not as significant as the improvement achieved by incorporating the LSceneLLM module, which validates the efficiency of our approach.

\para
\noindent\textbf{Dense Vision Token Selection Strategy.}
We conducted ablation experiments to verify that the attention map in the self-attention module reflects the visual information the model focuses on when answering questions. As shown in Tab.~\ref{abl_strategy}, the selection strategy based on attention weight outperforms the random selection strategy, demonstrating that the information about the regions that the model is currently focusing on aids in understanding the scene, while the random selection strategy provides little benefit to the model.

\section{More Scene Understanding Results on ScanNet}
\label{appendix:more_scannet_resules}
We also test our method on scene caption, embodied planning, and embodied qa, these datasets are sourced from the ScanNet part of 3D-LLM~\cite{hong20233d} and organized by Ll3da~\cite{chen2024ll3da}. Embodied QA requires the model to answer questions from the perspective of an agent, considering the agent's position and orientation within the environment. All of these tasks demand the model to have a holistic understanding of the entire scene. As shown in Tab.~\ref{tab:other_results}, our method outperforms the current state-of-the-art approaches on most metrics, demonstrating that the proposed approach not only captures fine-grained details in the scene but also achieves an accurate overall understanding of the entire scene.

\section{More Attention Visualization of LSceneLLM on XR-QA}
\label{appendix:viz}
We provide more attention map visualization results when LSceneLLM deals with different instructions on XR-QA. Experiment results show that our proposed method can accurately locate the task-relevant visual features using adaptive visual preferences from LLM.

\section{Computational Complexity Analysis}
\label{appendix:flops}
We analyze the computational complexity of the Ll3da~\cite{chen2024ll3da}, Leo~\cite{huang2023embodied}, and our method when faced with the same scene and identical text input. In Leo, the number of visual tokens corresponds to the number of objects in the scene, while both the Ll3da and our method use a fixed number of visual tokens when dealing with scenes of varying sizes, as shown in Tab.~\ref{tab:flops}. The computational complexity of our method is situated between the two baseline methods. As the scene size increases further, LSceneLLM can maintain a constant computational complexity while preserving the scene details. In contrast, the computational complexity of the object-centric method increases with the growing number of objects in the scene.

\end{CJK*}

\end{document}